# Retrofitting Concept Vector Representations of Medical Concepts to Improve Estimates of Semantic Similarity and Relatedness

Zhiguo Yu[a], Byron C. Wallace[b], Todd Johnson[a], Trevor Cohen[a]

[a] *The University of Texas School of Biomedical Informatics at Houston, Houston, Texas, USA,*
[b] *College of Computer and Information Science, Northeastern University, Boston, Massachusetts, USA,*

**Abstract**

*Estimation of semantic similarity and relatedness between biomedical concepts has utility for many informatics applications. Automated methods fall into two categories: methods based on distributional statistics drawn from text corpora, and methods using the structure of existing knowledge resources. Methods in the former category disregard taxonomic structure, while those in the latter fail to consider semantically relevant empirical information. In this paper, we present a method that retrofits distributional context vector representations of biomedical concepts using structural information from the UMLS Metathesaurus, such that the similarity between vector representations of linked concepts is augmented. We evaluated it on the UMNSRS benchmark. Our results demonstrate that retrofitting of concept vector representations leads to better correlation with human raters for both similarity and relatedness, surpassing the best results reported to date. They also demonstrate a clear improvement in performance on this reference standard for retrofitted vector representations, as compared to those without retrofitting.*

## Introduction

Incorporation of semantically related terms and concepts can improve the retrieval [1; 2] and clustering [3] of biomedical documents; enhance literature-based discovery [4; 5]; and support the development of biomedical terminologies and ontologies [6]. However, automated estimation of the semantic relatedness between medical terms in a manner consistent with human judgment remains a challenge in the biomedical domain. Many existing semantic relatedness measures leverage the structure of an ontology or taxonomy (e.g. WordNet, the Unified Medical Language System (UMLS), or the Medical Subject Headings (MeSH)) to calculate, for example, the shortest path between concept nodes [7-9]. Alternatively, vector representations derived from distributional statistics drawn from a corpus of text can be used to calculate the relatedness between concepts [7; 10]. Other corpus-based methods use information content (IC) to estimate the semantic relatedness between two concepts, from the probability of these concepts co-occurring [9; 11; 12]. This raises the question of whether knowledge- or corpus-based metrics are most consistent with human judgment.

In 2012, Garla and Brant [13] evaluated a wide range of lexical semantic measures, including both knowledge-based approaches leveraging the structure of an ontology or taxonomy [7; 14; 15] and distributional (corpus-based) approaches relying on co-occurrence statistics to estimate relatedness between concepts [16; 17]. This systematic investigation used several publicly available benchmarks. The most comprehensive of these is the University of Minnesota Semantic Relatedness Standards (UMNSRS), which contains the largest number and diversity of medical term pairs of any reference standard to date [18]. Medical terms in the set have been mapped to Concept Unique Identifiers (CUIs) in the UMLS, and term pairs have been annotated by human raters for similarity (e.g. *Lipitor* and *Zocor* are similar) and relatedness (e.g. *Diabetes* and *Insulin* are related). The best Spearman rank correlation for relatedness and similarity on this benchmark reported in [13] are 0.39 and 0.46 respectively.

Neural network based models that are trained to predict neighboring terms to observed terms, such as the architectures implemented by the *word2vec* package [19], have gained popularity as a way to obtain distributional vector representations of terms. Vectors induced in this way have been shown to effectively capture analogical relationships between words [20], and under optimized hyperparameter settings these models have been shown to achieve better correlation with human judgment than prior distributional models such as Pointwise Mutual Information (PMI) and Latent Semantic Analysis (LSA) on some word similarity and analogy reference datasets [21; 22]. However, embedding models are trained on terms, not concepts. In 2014 De Vine [23] and his colleagues demonstrated that word embedding models trained on sequences of UMLS concepts (rather than sequences of terms) outperformed established corpus-based approaches such as Random Indexing [24] and LSA [25].

In 2014 Sajadi et al. reported that a graph-based approach (HITS-sim) leveraging Wikipedia as a network outperformed *word2vec* trained on the OHSUMED corpus for the UMNSRS benchmark, with Spearman rank correlations of 0.51 and 0.58 for semantic relatedness and similarity respectively [26]. Most recently, Pakhomov et al. [27] performed an evaluation of *word2vec* trained on text corpora in different domains - Clinical Notes, PubMed Central (PMC), and Wikipedia - and achieved higher correlations of 0.58 and 0.62 for semantic relatedness and similarity respectively, which are the best results reported to date on the UMNSRS benchmark.

However, while vector representations produced by neural word embedding models are semantically informative, they disregard the potentially valuable information contained in semantic lexicons such as WordNet, FrameNet, and the Paraphrase Database. In 2015, Faruqui *et al.* developed a 'retrofitting' method that addresses this limitation by incorporating information from such semantic lexicons into word vector representations, such that semantically linked words will have similar vector representations [28]. In our previous work, we have tested this approach as a way to improve measures of semantic relatedness between MeSH terms using information from the MeSH taxonomic structure [29]. While retrofitted word vectors resulted in higher correlation with physician judgments, the reference set utilized was the MiniMayoSRS benchmark [7], which is a relatively small dataset (29 medical concept pairs). Furthermore, we did not apply neural word embeddings, which have been shown to outperform prior distributional models on this task.

In this paper, we extend our previous 'retrofitting' work in the following ways: (1) We use one of *word2vec*'s models to construct vector representations; (2) For construction of vector representations of UMLS concepts, we follow the approach described in [23] and train our model on sequences of UMLS medical concepts extracted from all of MEDLINE's titles and abstracts; (3) We evaluate our approach with a more extensive reference standard, the UMNSRS benchmark. Our results show that our method achieves higher correlation with human ratings for relatedness and similarity than the best results reported so far on UMNSRS benchmark [27].

## Methods

### Reference Standard

We used the University of Minnesota Semantic Relatedness Standard (UMNSRS) as our evaluation data [18]. This dataset consists of over 550 pairs of medical terms. Each term has been mapped to a CUI in the UMLS. Each pair of terms was assessed by 4 medical residents and scored with respect to the degree to which the terms were similar or related to each other, using a continuous scale. There are two subsets in UMNSRS - UMNSRS-Similarity and UMNSRS-Relatedness. UMNSRS-Similarity contains 566 pairs of terms rated by 4 medical residents. UMNSRS-Relatedness contains 587 pairs rated by 4 different medical residents. Each dataset can also be divided into 6 semantic categories: DISORDER-DISORDER, SYMPTOM-SYMPTOM, DISORDER-DRUG, DISORDER-SYMPTOM, DRUG-DRUG, and SYMPTOM-DRUG pairs.

In Pakhomov al et.'s evaluation, they modified the UMNSRS dataset to retain only those medical terms that appear in all of the three corpora that they used (Clinical Notes, PubMed Central articles, and Wikipedia). This reduced the number of pairs from 566 to 449 pairs in UMNSRS-Similarity, and from 588 to 458 pairs in UMNSRS-Relatedness.

In our evaluation, we use both the entire UMNSRS dataset and the modified UMNSRS dataset used by Pakhomov et al. For the full dataset, 526 of 566 pairs in UMNSRS-Similarity and 543 of 588 pairs in UMNSRS-Relatedness were found in our pre-processed corpus. For the modified dataset, this corpus contains 418 of 449 pairs for UMNSRS-Similarity and 427 of 458 pairs for UMNSRS-Relatedness.

### Semantic Lexicon from UMLS

The Unified Medical Language System is a repository of biomedical vocabularies developed by the US National Library of Medicine. It contains three components: the Metathesaurus; a Semantic Network, and the Specialist Lexicon (lexical information and tools for natural language processing). The Metathesaurus forms the base of the UMLS and comprises over 1 million biomedical concepts. It is organized by concept, and each concept has specific attributes defining its meaning and its links to corresponding concept names in the various source vocabularies [30]. In this work, we only used the UMLS Metathesaurus' "related concepts" file. This file contains all pair-wise relationships between concepts (or "atoms") known to the Metathesaurus. Table 1 displays different relationships and their descriptions.

*Table 1– Categories of relationships and their descriptions*

| Relationship | Description |
|---|---|
| AQ | has allowed qualifier |
| | *e.g. Myopathy AQ prevention & control* |
| CHD | has child relationship |
| | *e.g. Anemia CHD Mild anemia* |
| PAR | has parent relationship |
| | *e.g. Asthma PAR Bronchial Disease* |
| QB | can be qualified by |
| RB | has a broader relationship with |
| | *e.g. Angina RB Pain* |
| RL | the relationship is similar or "alike." |
| RN | has a narrower relationship with |
| | *e.g. Hernias RN Hernia, Paraesophageal* |
| RO | has other relationship |
| | *e.g. Ciprofloxacin RO Cipro 250 MG Tablet* |
| RQ | related and possibly synonymous. |
| | *e.g. Asthma RQ Wheezing* |
| RU | Related, unspecified |
| SIB | has sibling relationship |
| | *e.g. Acne SIB Skin Cancer* |
| SY | the source asserted synonymy |
| | *e.g. Diarrhea SY Dysentery* |
| XR | not related, no mapping |

For each concept in the evaluation dataset, we collected all related concepts within a one-step relationship from this related concepts file. For example, if A is our target concept and we have relationships A CHD B and B CHD C, only B will be considered as a semantic lexicon candidate for A.

### Concept-Based Word Embedding Model

To prepare the background corpus for the word embedding model, we downloaded all of the citations (titles and abstracts) in PubMed published before 2016. We then ran SemRep [31], which uses MetaMap [32] for concept extraction and normalization, on each citation's title and abstract to obtain a sequence of concept unique identifiers (CUI). In other words, following De Vine et al. [23], each sentence in this corpus is replaced by a sequence of CUIs, indicating the order in which concepts were encountered in the text.

To train this word embedding model, we used the *word2vec* implementation in Gensim, a Python package [33] to generate a 'concept embedding' for each CUI in our pre-processed corpus. We followed [27] in using the continuous bag-of-words (CBOW) model for word embedding training. The window size was set to 20, and the dimensionality of feature vectors was set at 200. We ignored all CUIs with a total frequency lower than 5.

### Retrofitting Word Vector to Semantic Lexicons

Vector space word representations are a critical component of many modern natural language processing systems. Currently it is common practice to represent words using corpus-derived dense high-dimensional vectors. However, this fails to take into account relational structures that have been explicitly encoded into semantic lexicons. Retrofitting is a simple and effective method to improve word vectors using word relationship knowledge encoded in semantic lexicons. It is used as a post-processing step to improve vector quality [28].

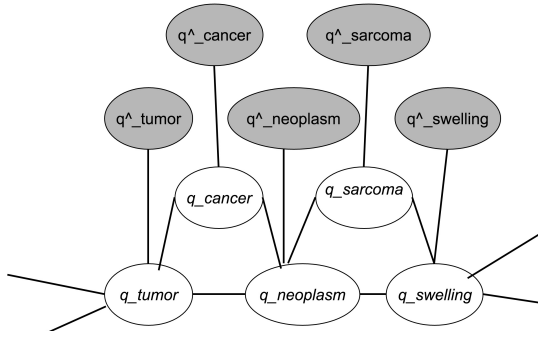

*Figure 1– Word graph with edges between related words, observed (gray node), inferred (white node)*

Figure 1 shows a small word graph example with edges connecting semantically related words. The words, *cancer*, *tumor*, *neoplasm*, *sarcoma*, and *swelling*, are similar words to each other, as defined in a lexical knowledge resource. Grey nodes represent observed word vectors built from the corpus. White nodes represent inferred word vectors, waiting to be retrofitted. The edge between each pair of white nodes means they represent related words (according to some knowledge source). The inferred word vector (e.g., q_tumor) is expected to be close to both its original (pre-retrofitting) estimated word vector (i.e., q^_tumor) and the retrofitted vector of its semantic neighbors (e.g., q_cancer and q_neoplasm). The objective is to minimize the following:

$$\psi(Q) = \sum_{i=1}^{n} [\alpha_i \parallel q_i - \hat{q_i} \parallel^2 + \sum_{(i,j) \in E} \beta_{ij} \parallel q_i - q_j \parallel^2 ]$$

where $\alpha$ and $\beta$ are hyperparameters that control the relative strengths of corpus- and lexically-derived associations, $Q$ represents the retrofitted vectors, and $(i,j) \in E$ means there is an edge between node $q_i$ and $q_j$. $\Psi$ is convex in $Q$. An efficient iterative updating method is used to solve the convex objective. First, retrofitted vectors in $Q$ are initialized to be equal to the empirically estimated vectors. The next step is to take the first derivative of $\Psi$ with respect to the $q_i$ vector and use the following to update it online.

$$q_i = \frac{\sum_{j:(i,j) \in E} \beta_{ij} q_j + \alpha_i \hat{q_i}}{\sum_{j:(i,j) \in E} \beta_{ij} + \alpha_i}$$

In practice, it takes approximately 10 iterations to converge to the difference in Euclidean distance of adjacent nodes of less than 0.01. We used the authors' implementation of this algorithm [28].

### Evaluation Measures

In the evaluation, we tested different semantic lexicons (based on the categories of relationships described in Table 1) with the 'retrofitting' method to improve the vector quality of each concept. For each term pair in the test dataset, we extracted concept vectors and computed the cosine similarity between them using the following equation:

$$\cos(\theta) = \frac{A \cdot B}{\parallel A \parallel \parallel B \parallel} = \frac{\sum_{i=1}^{N} A_i B_i}{\sqrt{\sum_{i=1}^{N} A_i^2} \sqrt{\sum_{i=1}^{N} B_i^2}}$$

Where $A_i$ and $B_i$ are components of vector $A$ and $B$ respectively, and $N$ is the length of vector. The cosine scores computed for each pair in the test dataset were then compared to the mean of the human similarity and relatedness judgments for each pair, using Spearman rank correlation. We also tested our method on different subsets of the UMNSRS dataset consisting of pairs of different semantic types. The baselines we used for comparison are the results reported by Pakhomov *et al.* in 2016 [27].

### Results

#### Comparisons with different lexicons

The results of these experiments are shown in Table 2, which shows results after retrofitting for all relationship types in Table 1.. Given differences in vocabulary across corpora, we cannot compare the identical set of pairs used by Pakhomov *et al.* Nonetheless, our CUI-based vector representations based without retrofitting ("No Retrofitting") perform slightly better than the results reported by Pakhomov *et al.* on both full and modified UMNSRS sets. Retrofitting with RO relationships results in best performance for semantic similarity, with correlations of 0.683 and 0.673 for the full and modified datasets, respectively. For UMNSRS-Relatedness, using RQ relationships achieves best performance with correlations of 0.609 and 0.621 for the full and modified dataset, respectively.

*Table 2-Comparison of Spearman rank correlations between human raters and our method using different lexicons*

| Pakhomov al et. | 0.62 (n=449) | | 0.58 (n=458) | |
|---|---|---|---|---|
| UMNSRS | Similarity | | Relatedness | |
| Test Subsets | Full (n=526) | Modified (n=418) | Full (n=543) | Modified (n=427) |
| No Retrofitting | 0.639 | 0.628 | 0.585 | 0.594 |
| AQ | 0.574 | 0.552 | 0.527 | 0.525 |
| SIB | 0.601 | 0.585 | 0.530 | 0.535 |
| PAR | 0.632 | 0.618 | 0.561 | 0.562 |
| RB | 0.636 | 0.624 | 0.586 | 0.593 |
| RL | 0.639 | 0.628 | 0.585 | 0.594 |
| RU | 0.639 | 0.628 | 0.585 | 0.594 |
| QB | 0.639 | 0.628 | 0.585 | 0.594 |
| XR | 0.639 | 0.628 | 0.585 | 0.594 |
| CHD | 0.642 | 0.632 | 0.588 | 0.595 |
| SY | 0.654 | 0.644 | 0.599 | 0.610 |
| RQ | 0.657 | 0.655 | **0.609** | **0.621** |
| RN | 0.664 | 0.656 | 0.600 | 0.608 |
| RO | **0.683** | **0.673** | 0.604 | 0.613 |

Only lexical information concerning *CHD, SY, RQ, RN,* and *RO* relationships improved the performance of concept vector representations. Table 3 presents the performance of our retrofitting method using different combinations of productive relationships on the test dataset. Combining *RN* and *RO* relationships resulted in the best performance of 0.689 and 0.681 for the full and modified UMNSRS-Similarity datasets.

*Table 3-Comparison of Spearman rank correlations between human raters and our method using lexicons combinations*

| Pakhomov al et. | 0.62 (n=449) | | 0.58 (n=458) | |
|---|---|---|---|---|
| | UMNSRS-Similarity | | UMNSRS-Relatedness | |
| Lexicons Combinations | Full (n=526) | Modified (n=418) | Full (n=543) | Modified (n=427) |
| No Lexicons | 0.639 | 0.628 | 0.585 | 0.593 |
| CHD+SY | 0.651 | 0.643 | 0.596 | 0.605 |
| RQ+RO | 0.686 | 0.679 | 0.616 | 0.627 |
| RN+RQ | 0.667 | 0.662 | 0.607 | 0.617 |
| RN+RO | **0.689** | **0.681** | 0.619 | 0.630 |
| RN+RO+RQ | 0.687 | 0.681 | 0.622 | 0.634 |
| SY+RN+RO+RQ | 0.686 | 0.680 | 0.623 | 0.634 |
| CHD+SY+RN+RO+RQ | 0.686 | 0.680 | **0.624** | **0.635** |

For UMNSRS-Relatedness, lexicons with all five productive relationships attained the highest correlations of 0.624 and

0.635 for the full and modified datasets respectively. Furthermore, any lexicons including *RO* relationship have similar performance for UMNSRS-Similarity and any lexicons including RQ obtain similar correlation scores for UMNSRS-Relatedness.

**Comparison across pairs of different semantic types**

From Table 3, we can see that lexicons containing *RN+RO* and *CHD+SY+RN+RO+RQ* achieved the best performances for UMNSRS-Similarity and UMNSRS-Relatedness respectively. Hence, we just used these two lexicons in the comparison of Spearman rank correlations between human raters and our method in different subsets of pairs grouped by semantic types. Table 4 and Table 5 present performances of comparisons for UMNSRS-Similarity and UMNSRS-Relatedness. As shown in Table 4, the lexicon from *RN* and *RO* relationships achieves the best correlation performance in 4 of 6 groups and lexicon from *CHD, SY, RN, RO,* and *RQ* relationships obtain the highest correlation score in symptom-symptom pairs. However, Pakhomov *et al.* retain the best performance for disorder-disorder (Di-Di) pairs, using PMC.

*Table 4-Comparison of Spearman rank correlations between human raters estimates of similarity and our method in different subsets of pairs grouped by semantic types (Di-disorder, S-symptom, Dr-drug)*

| UMNSRS-Similarity | Pakhomov et al. | RN+RO | | CHD+SY+RN+RO+RQ | |
|---|---|---|---|---|---|
| | Highest | Mod | Full | Mod | Full |
| All Pairs | 0.62 | 0.681 | **0.689** | 0.680 | 0.686 |
| Di-Di | **0.74** | 0.715 | 0.72 | 0.723 | 0.726 |
| S-S | 0.56 | 0.625 | 0.668 | 0.635 | **0.670** |
| Dr-Dr | 0.77 | **0.841** | 0.749 | 0.840 | 0.748 |
| Di-S | 0.49 | 0.703 | **0.720** | 0.699 | 0.717 |
| Di-Dr | 0.69 | 0.686 | **0.710** | 0.682 | 0.708 |
| S-Dr | 0.51 | 0.484 | **0.552** | 0.476 | 0.546 |

As shown in Table 5, the lexicon containing *CHD, SY, RN, RO,* and *RQ* relationships resulted in the highest correlation with human raters in 4 of 6 groups for UMNSRS-Relatedness dataset. Pakhomov *et al.* retained the best performance in disorder-drug (Di-Dr) and symptom-drug (S-Dr), achieved using embeddings trained on clinical notes [27].

*Table 5- Comparison of Spearman rank correlations between human raters estimates of relatedness and our method in different subsets of pairs grouped by semantic types (Di-disorder, S-symptom, Dr-drug)*

| UMNSRS-Relatedness | Pakhomov et al. | RN+RO | | CHD+SY+RN+RO+RQ | |
|---|---|---|---|---|---|
| | | Mod | Full | Mod | Full |
| All Pairs | 0.58 | 0.630 | 0.619 | **0.635** | 0.624 |
| Di-Di | 0.59 | 0.589 | 0.628 | 0.593 | **0.629** |
| S-S | 0.64 | 0.692 | 0.706 | 0.724 | **0.730** |
| Dr-Dr | 0.73 | 0.734 | 0.571 | **0.736** | 0.572 |
| Di-S | 0.42 | 0.562 | 0.594 | 0.569 | **0.603** |
| Di-Dr | **0.63** | 0.564 | 0.607 | 0.565 | 0.611 |
| S-Dr | **0.59** | 0.479 | 0.519 | 0.482 | 0.523 |

## Discussion

In this study, we used a method for retrofitting of word embeddings to improve semantic similarity and relatedness measures by incorporating structural information from the UMLS. We evaluated our approach on both the full UMNSRS dataset and the modified subset used in [27]. Vector representations trained on sequences of CUIs (without retrofitting) resulted in comparable performance (with slight improvements) to those based on sequences of terms. After applying retrofitting on CUI vector represents using selected UMLS relationship types, we see clear improvements on both the full and modified dataset, compared to the CUI vectors without retrofitting. In comparison with the best results previously reported on the UMNSRS benchmark ([27] - 0.62 for similarity and 0.58 for relatedness), we obtain better correlation with human raters on both similarity and relatedness (0.689 for similarity and 0.624 for relatedness on the full UMNSRS dataset and 0.681 for similarity and 0.635 for relatedness on the modified version). However, as our results concern a subset of the modified set only, further evaluation on matching sets is required to show this conclusively. Our codes and word embeddings are available at

(*https://github.com/Sssssssstanley/Retrofitting-Concept-Vector-Representations-of-Medical-Concepts*).

However, our results also show that external linkage information should be carefully chosen. For example, using *AQ, SIB, PAR,* and *RB* relationships resulted in *worse* correlation with human judgment than the original concept vectors (without retrofitting). This suggests that these relationship types are too permissive to align with human evaluation. Incorporating other relationships, such as *RB, RL, RU,* and *XR*, had no effect on the results. The reason for this is that no CUIs connected to CUIs in the evaluation set using these relationships. *CHD, SY, RQ, RN,* and *RO* clearly have positive effects on the quality of the vector representations. *RO* has the largest positive effect on the Similarity dataset, and *RQ* improves the vector presentation the most on the Relatedness dataset. The description of *RO* is 'has a relationship other than synonymous, narrower, or broader.' For example, *Ciprofloxacin* and *Cipro 250 MG Oral Tablet* are linked by *RO*. These are the same drug with different dosages, so retrofitting would enhance similarity between vectors for concepts representing the same drug. The description of *RQ* is 'related and possibly synonymous'. "Relatedness" is a general notion that encompasses similarity, and maps well to this relationship type. Hence, it seems reasonable that incorporating this relation would achieve the best correlation with human raters on UMNSRS-Relatedness dataset.

As noted in [27] the correlations in the 0.5~0.6 range reported for the UMNSRS benchmark are in the same range as the intra-class correlation coefficients used to measure agreement between human annotators for this set, and so may constitute the ceiling for performance that can be measured using this benchmark. However, our results are clearly over this range. What we reported are correlations with the mean rating, which may be more readily approximated than the ratings of a single rater. In the future, we will conduct further analysis on interpreting our results in relation to the inter-rater agreement intra-class correlations for different categories of term pairs.

## Conclusions

In this paper, we introduced a hybrid method for generating semantic vector representations of UMLS concepts, by leveraging both distributional statistics and linkage information from an ontology or taxonomy (such as the UMLS). This method achieved better performance on the UMNSRS benchmark than neural word embeddings alone, with the best results reported for this evaluation to date. Any application using concept vector representations could potentially benefit from the additional structural information encoding using this retrofitting approach. In the future, we will continue to evaluate the utility of retrofitting method for downstream tasks (such as word-sense disambiguation and information retrieval).

## Acknowledgements

This work was supported by the UTHealth Innovation for Cancer Prevention Research Training Program Predoctoral Fellowship (Cancer Prevention and Research Institute of Texas (CPRIT) grant # RP160015), and supported in part by National Library of Medicine R01LM011563. The content is solely the responsibility of the authors and does not necessarily represent the official views of the CPRIT or NLM.

**Address for correspondence**
*Trevor Cohen, MBChB, PhD*
*7000 Fannin St, Suite 600, Houston, TX, 77030*
*Email : Trevor.Cohen@uth.tmc.edu*
*Phone : 713.486.3675  Fax : 713.486.0117*